\documentclass[12pt,letterpaper]{article}

\usepackage[pagenumbers]{cvpr} 

\usepackage{graphicx}
\usepackage{amsmath}
\usepackage{amssymb}
\usepackage{booktabs}
\usepackage{algorithm}   
\usepackage{algpseudocode}
\usepackage{siunitx}
\newcommand{\grad}{\ensuremath{\nabla}}

\usepackage[pagebackref,breaklinks,colorlinks]{hyperref}

\usepackage[capitalize]{cleveref}
\crefname{section}{Sec.}{Secs.}
\Crefname{section}{Section}{Sections}
\Crefname{table}{Table}{Tables}
\crefname{table}{Tab.}{Tabs.}

\begin{document}

\title{Semi-Automated Segmentation of Geoscientific Data Using Superpixels}

\author{Conrad Koziol\\
Computational Geosciences\\
Vancouver, Canada\\
{\tt\small conrad@compgeoinc.com}
\and
Eldad Haber\\
The University of British Columbia \\
Vancouver, Canada\\
{\tt\small ehaber@eoas.ubc.ca}
}
\maketitle

\begin{abstract}
Geological processes determine the distribution of resources such as critical minerals, water, and geothermal energy. However, direct observation of geology is often prevented by surface cover such as overburden or vegetation. In such cases, remote and in-situ surveys are frequently conducted to collect physical measurements of the earth indicative of the geology. Developing a geological segmentation based on these measurements is challenging since individual datasets can differ in properties (e.g. units, dynamic ranges, textures) and because the data does not uniquely constrain the geology. Further, as the number of datasets grows the information to constrain geology increases while simultaneously becoming harder to make sense of. Inspired by the concept of superpixels, we propose a deep-learning based approach to segment rasterized survey data into regions with similar characteristics. We demonstrate its use for semi-automated geoscientific mapping with datasets arising from independent sensors and with diverse properties. In addition, we introduce a new loss function for superpixels including a novel regularization parameter penalizing image segmentation with non-connected component superpixels. This improves integration of prior knowledge by allowing better control over the number of superpixels generated.
\end{abstract}

\section{Introduction}
\label{sec:intro}

Understanding the geology of a domain is one of the fundamental tasks in the earth sciences. Geological processes are a controlling factor in the earth's evolution, and determine the distribution of resources such as critical minerals, groundwater hydrology, and geothermal energy. However, geology often cannot be observed directly due to the presence of overburden or vegetation. As an alternative, physical properties of the earth indicative of the geology will frequently be measured by conducting remote 
surveys such as magnetic, gravity, and airborne electromagnetic. This data is often interpreted with a sparse set of geological observations in an expert-driven process manually delineating a geological map.
The basic premise of this interpretation is that different geological units have different physical properties and therefore, geophysical observations at scale depend on the geology.

Geological features are not uniquely determined by survey data. This is because the data reflects a certain physical property rather than a geological process.
Therefore, an expert-judgement is relied on for interpretation. This has parallels to medical imaging where radiologists interpret medical images, and contrasts with classic computer vision problems, where grayscale or RGB data is sufficient for instance segmentation. A spectrum of realizations of a geological map can be equally valid, with a subjective selection of a singular output. In many cases it is impossible to know which interpretation is more feasible which makes computerized learning the task of geological segmentation difficult.

Multiple heterogeneous data sources containing independent information are often integrated in geological mapping. Data sources include geophysical surveys, geochemical soil sampling, radiometrics, and hyperspectral imagery. Because data sources measure independent physical properties, images appear visually distinct and can have significantly different features. Increasing numbers of data layers provides better information to constrain geology while simultaneously becoming harder to make sense of.

In this paper, we consider the problem of automated segmentation of a geological domain using rasterized survey data. Inspired by the concept of superpixels, we aim to a the domain into regions with similar geological characteristics. We advance on previous efforts by taking a deep-learning approach and considering datasets derived from independent sensors with measurements of disparate characteristics. Further, we introduce a new regularization function to facilitate human-in-the-loop control over the number of superpixels generated by soft-superpixel assignment methods.

We approach the problem as an single-shot unsupervised learning image segmentation problem. Our main contributions are:
\begin{itemize}
  \item Demonstrate the use of superpixels for semi-automating geoscientific mapping
  \item Propose a new loss function for superpixel mapping, including a novel regularization function penalizing non-connected component superpixels.
\end{itemize}

The rest of the paper is organized as follows.
In \cref{sec2} we explore related work and discuss the differences between our work to other known techniques. In \cref{sec3} we describe the superpixel method and its training on data. In \cref{sec4}
we demonstrate the utility of our methodology by an application to the Yalgin Craton in Western Australia
and we summarize the paper in \cref{sec5}.

\section{Related Works}
\label{sec2}

Previous applications of superpixels for geological understanding on a regional scale have focused on spectral imagery. Superpixel segmentation of hyperspectral imagery has been used to determine mineral spectra end members \cite{gilmore2011superpixel,thompson2010superpixel} and to select bands for lithological discrimination \cite{tan2020hyperspectral}. The work presented in \cite{vasuki2017interactive} develop a semi-automated tool for lithological boundary detection in optical images based on user input and an iterative algorithm merging superpixels. An automated approach to lithological segmentation of optical images is developed in \cite{sang2020intelligent} based on merging independent segmentations from simple linear iterative clustering (SLIC) \cite{achanta2012slic} and a convolutional neural network (CNN).

Numerous image segmentation algorithms have been applied to geological data for detecting structures, lithological units, and alteration \cite{shirmard2022review}. Terracing was an early method used to delineate structures in potential field data \cite{cordell1989terracing, phillips1992terrace}. CNNs have been applied to predict preliminary geological form airborne magentic data \cite{cedou2022preliminary}, spectral imagery \cite{sang2020intelligent, latifovic2018assessment}, and geochemical data \cite{wang2022geological}. Notably, self-organizing maps with k-means clustering has been used by \cite{carter2021predictive, carneiro2012semiautomated} to predict geological units from multiple geophysical datasets.

This work to our knowledge is the first to propose using superpixels to segment a set of geoscientific datasets. In contrast to previous superpixel applications in the geosciences, which focus on spectral images, we consider data from multiple sensors. The data considered not only have different units and ranges, but are independent of each other, and do not have the same qualitative features (e.g. patterns, textures, edges). In this context superpixels can be thought of as a method for data fusion. By demonstrating that superpixels are an effective tool for segmentation, we provide an alternative approach to self-organizing maps with k-means clustering based on modern deep-learning.

\begin{figure*}[ht!]
\includegraphics[width=\textwidth]{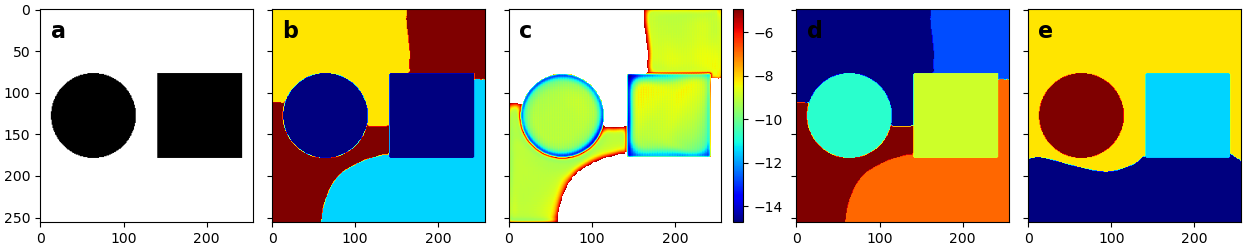}
\centering
\caption{The impact of $\mathbf{R}_{cc}$ demonstrated on a toy problem. a) Toy dataset with background values of zero and foreground values of 1. b) Image segmentation ($\mathbf{S}$) with four output channels ($N$) and the coefficient of $\mathbf{R}_{cc}$ set to zero. Notice that the segmentation is discontinuous. Both the circle and square are assigned to the same segment, as are the segments in the top right and bottom left corners. c) $-\| \nabla \mathbf{R}_{cc} \|_2$ pixelwise with respect to the logits. d) Superpixel segmentation ($\mathbf{S}'$) of the output shown in panel (b). Notice there are 6 superpixels. e) Image segmentation  leveraging $\mathbf{R}_{cc}$. Note that segments are continuous, and the number of superpixels corresponds to the number of output channels.}
\label{fig:stoy}
\end{figure*}

\section{Methods}
\label{sec3}

We take an unsupervised single-shot image-segmentation approach to generating superpixels. The description of our methodology follows in three parts. We begin with a preliminary section introducing key notation and standard post-processing applied to neural network outputs to generate superpixels. Next we discuss the core of our methodology which is the objective function we minimize. Finally we detail the neural network architecture and training procedure we use.

\subsection{Preliminaries}

The input image is denoted by $\mathbf{I} \in \mathbb{R}^{C\times H\times W}$, where $C$ is the number of channels, $H$ is the height of the image, and $W$ is the width of the image. The height and width of the image correspond to the spatial discretization of the domain, while the number of channels corresponds to the available datasets.
In our context datasets can be rather different, such as having different physical units and corresponding to different physical properties.

The neural network is specified as $f\left(\mathbf{I}, \theta \right): \mathbb{R}^{C\times H\times W} \times \mathbb{R}^n \to \mathbb{R}^{N\times H\times W}$, where $N$ is the target number of superpixels. Note that each superpixel is treated as a different channel or class. The output of the neural network is used to generate two probabilistic pixelwise image segmentations. These appear in the objective function and motivated further in that context:

\begin{equation}
\mathbf{P} = \textsf{softmax}_n\left( f\left(\mathbf{I}, \theta \right)\right)
\end{equation}
\begin{equation}
\mathbf{P}^{s} = \textsf{sparsemax}_n\left( f\left(\mathbf{I}, \theta \right) \right)
\end{equation}

where $\mathbf{P} \in \mathbb{R}^{N\times H\times W}$, $\mathbf{P}^{s} \in \mathbb{R}^{N\times H\times W}$, and the indices $n,i,j$ are over the $N, H, W$ dimensions respectively. The $\textsf{sparsemax}_n\left(\cdot \right)$ operator is analogous to the softmax function but able to output sparse probabilities \cite{martins2016softmax}. It's defined as the euclidean projection of the input vector onto the probability simplex:
\begin{equation}
\textsf{sparsemax}\left(\mathbf{z} \right) := \textsf{argmin}_{\mathbf{p} \in \Delta^{K-1}} || \mathbf{p}  - \mathbf{z}  ||^2
\end{equation}

where $\Delta^{K-1}$ is the $(K-1)$-dimensional probability simplex. Additionally, the neural network output is used to calculate an image segmentation $\mathbf{S} \in \{0,1,..., N-1\}^{H\times W}$ consisting of $N$ classes:
\begin{equation}
\mathbf{S} = \textsf{argmax}_n\left( f\left(\mathbf{I}, \theta \right) \right)
\end{equation}

$\mathbf{S}$ is precursor to the superpixel map since there is no constraint that the image segments are continuous. Spatially independent regions within the image can be assigned the same classification, and hence are not superpixels by definition. This is akin to the difference between semantic and instance segmentation. 

Following \cite{achanta2012slic,suzuki2020superpixel, eliasof2022rethinking} we postprocess our superpixel segmentation to enforce pixel connectivity (using code available in SciPy \cite{2020SciPy-NMeth}). The primary aim of separating spatially discontinuous segments is achieved via a connected component algorithm. Additionally, segments can be either merged or split to enforce superpixel minimum and maximum size constraints. We denote this postprocessed image as $\mathbf{S}' \in \mathbb{R}^{H\times W}$, which is our image superpixel segmentation.

$\mathbf{S}'$ can be used to construct the following superpixelated representation of an image, which useful for visualizing single datalayers and RGB images:

\begin{equation}
\mathbf{I}_{ij}^h = \sum_n \mathbf{1}_n \left( \mathbf{S}_{ij}' \right) \frac{\sum_{ij} \mathbf{1}_n \left( \mathbf{S}_{ij}' \right) \mathbf{I}_{ij}}{\sum_{ij}  \mathbf{1}_n \left( \mathbf{S}_{ij}' \right)}
\end{equation}

The image $\mathbf{I}^h$ visualizes each superpixel with an average colour value over all individual pixels within the superpixel.

\subsection{Optimization Problem}

The core of our superpixel methodology is minimizing the following objective function:
\begin{equation}
\label{eq:loss_func}
\begin{split}
J(\mathbf{I},\theta) = c_1 \sum_{n} \left( \sum_{ij} \| \left( \mathbf{c}_{n}^s- \mathbf{I}_{nij} \right) \mathbf{P}_{nij}^{s} \|_{2}^{2} \right) + \\
c_2 \sum_{ij} \frac{1}{2} \| \grad \mathbf{P}^{s}_{ij}\|_{2}^{2} + 
c_3 \mathbf{R}_{clust}\left(\mathbf{P} \right) + c_4 \mathbf{R}_{cc}\left(\mathbf{P} \right)
\end{split}
\end{equation}

where $\mathbf{c}_{n}^s = \frac{\sum_{ij} \mathbf{P}_{nij}^{s} \mathbf{I}_{nij}}{\sum_{ij} \mathbf{P}_{nij}^{s}}$ is the average dataset values for an image segment, $\mathbf{R}_{clust}\left(\mathbf{P} \right)$ and $\mathbf{R}_{cc}\left(\mathbf{P} \right)$ are regularizers defined below, and $c_1 \to c_4$ are weighting parameters. 

The first term promotes an image segmentation grouping pixels with similar data values. Similarity is measured using the $\mathbf{L}^2$ norm, and is function of the difference between the average dataset values for a segment and pointwise dataset values. Multiplication by $\mathbf{P}_n^{s}$ weights the similarity by the probability of a pixel being assigned to the image segment, and is a differentiable approximation to integrating over the superpixel domains in $\mathbf{S}'$. We expect most of the image domain to have a low probability of belonging to a particular segment. The motivation to use $\mathbf{P}^{s}$ rather than $\mathbf{P}$ is therefore to use sparsity to attenuate the impact of a large number of low probability points.

The second term penalizes the length of boundaries of image segments, and is motivated as a prior that superpixels should have simple geometries. The term $\grad \mathbf{P}^{s}$ identifies locations where the probability of segmentation changes, and is used as a proxy for segment boundaries. We select $\mathbf{P}^{s}$ since we do not require full support over the segmentation probabilities.

\cite{kim2019mumford} previously define a loss function similar to the first two terms above for image segmentation (but not superpixels), inspired by the well studied Mumford-Shah Functional \cite{mumford1989optimal} in computer vision. The two differences with our formulation are that we use $\mathbf{P}^{s}$ rather than $\mathbf{P}$, and use an $L^2$ norm in the second term rather than $L^1$ norm.

$\mathbf{R}_{clust}$ is introduced in \cite{suzuki2020superpixel} and is based on the concept of regularized information maximization:

\begin{equation}
\begin{split}
\mathbf{R}_{clust} \left( \mathbf{P} \right)= \frac{1}{HW} \sum_{n} \sum_{ij} - \mathbf{P}_{nij} \log\left( \mathbf{P}_{nij} \right) + \\
\lambda \sum_n \mathbf{\hat{P}}_n \log \mathbf{\hat{P}}_n
\end{split}
\end{equation} 
 
where $\mathbf{\hat{P}}_n = \frac{1}{HW}\sum_{ij} \mathbf{P}_{nij}$, and $\lambda$ is a weighting factor. Minimizing the first term decreases the entropy of the pixelwise probability distribution of superpixel assignment, while minimizing the second term encourages a similar number of pixels to be assigned to each superpixel.

We introduce a new regularization term applicable to instance-segmentation type problems that penalizes spatial discontinuity. $\mathbf{R}_{cc}$ promotes image segments forming a single connected component in line with the definition of superpixels. When a one-to-one correspondence between image segments and superpixels exists to the first two terms in the objective function (\cref{eq:loss_func}) are calculated over individual superpixels rather than over arbitrary combinations. 

A practical benefit of this regularization term is to provide the user finer control over the number of superpixels. The number of superpixels can be made to be approximately the number of output channels ($N$). Without this term, the number of superpixels can be far greater than the number of image segments. \cref{fig:stoy} demonstrates the impact of $\mathbf{R}_{cc}$ on a toy example.

$\mathbf{R}_{cc}$ is defined as:

\begin{equation}
\mathbf{R}_{cc} = \sum_{n} 1\left( \Omega_n^r \neq \null \varnothing \right) \left( \sum_{ij \in \Omega_n^r} p_{ij} - \sum_{ij \in \Omega_n^d} \log p_{ij}  \right)
\end{equation} 

where $\Omega_n$ is the region segmented by the n\textsuperscript{th} superpixel (i.e. $1\left(\mathbf{S} = n \right)$), $\Omega_n^d$ is the largest connected component of that region, and $\Omega_n^r = \Omega_n - \Omega_n^d$ is the remaining area outside of the largest connected component. An algorithmic implementation of $\mathbf{R}_{cc}$ is shown in \cref{alg:rcc}.

\begin{algorithm}
\caption{Connected Component Regularization}\label{alg:cap}
\label{alg:rcc}
\begin{algorithmic}
\Procedure{$\mathbf{R}_{cc}$}{$f \gets f\left(\mathbf{I}, \theta \right)$}
\State $p \gets \textsf{softmax}\left( f \right)$
\State $s \gets \textsf{argmax}_n\left( f \right)$
\State $c \gets \textsf{connected\_components}\left( s \right)$
\State $m_{avers}, m_{attr} \gets \textsf{zeros\_like}\left( f \right)$
    \For{$n$ \texttt{in}  \textsf{unique}$\left( s \right)$}
    \State $v \gets c\left[s = n\right]$
        \If{$\textsf{length}\left(\textsf{unique}\left( v \right)\right) > 1$}
            \State $a,b \gets \textsf{unique}\left( v, \textsf{return\_counts} \right))$
            \State $a \gets a\left[ \textsf{argsort}\left( b, \textsf{descending}\right)\right]$
            \State $m_{aver}[n,:,:] \gets \left( s = n \right) \texttt{and} \left( cc \neq a[0] \right)$
            \State $m_{attr}[n,:,:] \gets \left( s = n \right) \texttt{and} \left( cc = a[0] \right)$
        \EndIf
    \EndFor
\State \algorithmicreturn{ $\textsf{mean} \left( m_{aver} p - m_{attr} \log \left( p \right) \right)$}
\EndProcedure
\end{algorithmic}
\end{algorithm}

At the core of $\mathbf{R}_{cc}$ is a non-differentiable connected component algorithm. We use this to determine the segments which split into multiple connected components. For each such segment, the dominant connected component is identified as having the greatest number of pixels. The pixel locations of the dominant superpixel are one region ($m_{attr} \in \left[0,1 \right]^{N\times H\times W}$ in \cref{alg:rcc}). The remaining non-dominant pixels are  grouped into another ($m_{aver} \in \left[0,1 \right]^{N\times H\times W}$  in \cref{alg:rcc}). The regularizer is a sum of terms which repels the non-dominant segmented areas away from their current classification and decreases the cross-entropy of dominant regions. The regularizer only impacts segmentations which do not form a single connected component. We find the regularizer to be relatively insensitive to the precise function of $p$ used (e.g. $p$ vs $\log{p}$).

\subsection{Implementation}

We use a UNet architecture \cite{ronneberger2015u} for our neural network $f\left(\mathbf{I}, \theta \right)$, on the basis of its effectiveness for segmentation problems and ability to scale with image dimensions. Individual blocks in the UNet are composed of sequences of convolution, ReLU, and Instance Norm. We fix the number of downsampling and upsampling blocks to four, with the number of channels increasing to 64 in the opening block and doubling with each block to reach a maximum of 512 in the bottleneck. Code is written using the PyTorch framework \cite{paszke2019pytorch}.

Based on experimentation, we find Adam \cite{kingma2014adam} with a learning rate of 1e-2 an effective optimizer. We found objective function weighting parameters of $c_1=1, c_2 =$1e-4$, c_3=100, c_4=50$, and $\lambda=2$ to produce good results, but vary parameters on the basis of the problem. Input images are normalized layerwise, and each dimension is resized to the nearest multiple of 16 to allow four downsampling operations in the UNet. 

\begin{figure}[ht!]
\includegraphics[width=5cm]{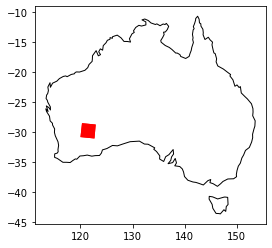}
\centering
\caption{Location of study domain in Australia. Coordinates in WGS84.}
\label{fig:extent}
\end{figure}

\begin{figure*}[ht!]
\includegraphics[width=\textwidth]{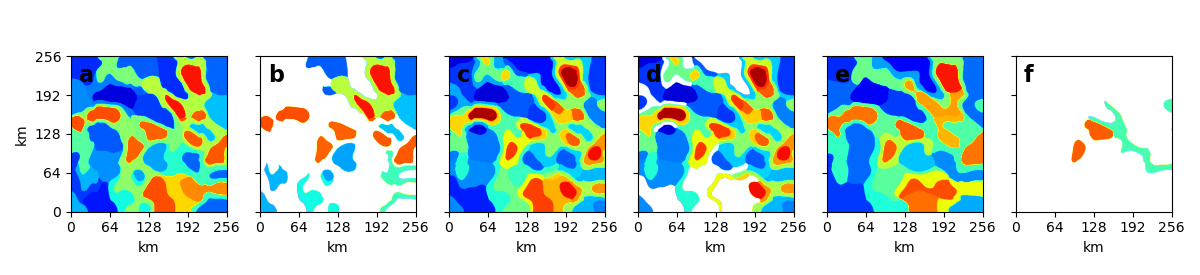}
\centering
\caption{The impact of $\mathbf{R}_{cc}$ demonstrated on the tau dataset. Panels display representation of the dataset ($\mathbf{I}^h$) created using pre-superpixel image segmentations $\mathbf{S}$ a) Using code from \cite{eliasof2022rethinking}. b) Visualization in (a) but masking single connected component segments. c) Our code without $\mathbf{R}_{cc}$. d) Analogous to (b). e) Our code with $\mathbf{R}_{cc}$. e) Analogous to (b). Observe there are far fewer non-contiguous segments and that they cover significantly less of the domain. }
\label{fig:regularization}
\end{figure*}

\section{Experiments} 
\label{sec4}

We apply superpixel methodology to a \SI{256}{\kilo\meter}  by \SI{256}{\kilo\meter} domain located in the Yalgin Craton region in Western Australia (\cref{fig:extent}). Three data layers available in this region relevant to mineral prospectivity mapping are used for superpixel generation: gravity (Complete Bouguer Anomaly \cite{lane20202019}), magnetic (Total Magnetic Intensity \cite{ga2020mag}), and tau (Time Constant from Airborne Electromagnetic Surveys \cite{leycooper2021}). Each datalayer is gridded at \SI{500}{\meter} resolution, and has dimensions of \SI{512}x\SI{512} pixels. We apply quantile normalization to the mag data due to its high dynamic range. In addition, we standardize all layers before inputing to the neural network.

\begin{figure*}[ht!]
\includegraphics[width=12cm]{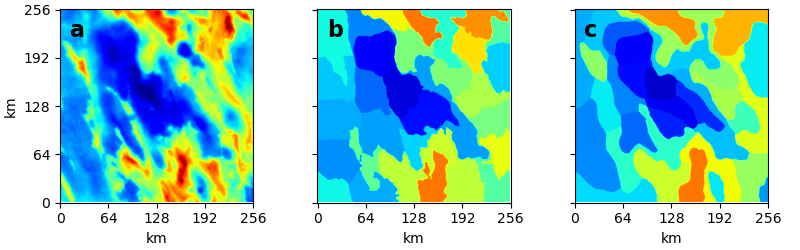}
\centering
\caption{Superpixel segmentation applied to gravity data. a) gravity data. b) SLIC superpixel image ($\mathbf{I}^h$). c) Our method superpixel image ($\mathbf{I}^h$).}
\label{fig:slic_comp}
\end{figure*}

As a first step, we demonstrate the impact of $\mathbf{R}_{cc}$ by generating a superpixel representation of the tau dataset with two unsupervised deep learning approaches: our method and that of \cite{eliasof2022rethinking}. We set the number of output channels (logits) for the networks to be 35. Results are plotted (\cref{fig:regularization}) before post-processing enforcing connectivity (i.e. $\mathbf{S}$). The output of the methodology of \cite{eliasof2022rethinking}  and our methodology without $\mathbf{R}_{cc}$ leads to 53 and 83 connected components respectively. Neglecting postprocessing, these would be the number of superpixels. Adding $\mathbf{R}_{cc}$ to our model decreases the number of connected components to 34. 

$\mathbf{R}_{cc}$ allows user-control over the number of superpixels generated. It's effect is to penalize image segments representing multiple superpixels, pushing the system to have no more than $N$ superpixels. The ability to constrain the number of superpixels is essential when the number of superpixels is selected based on prior knowledge such physical characteristics of the system, downstream task requirements, or user preference. 

We also contrast superpixels (\cref{fig:slic_comp}) based on the tau data using both our unsupervised deep learning approach and SLIC. Due to the numerous parameter choices in each model and subjectivity of the problem, the choice of algorithm can viewed as part of the modelling process. However, the advantages of deep learning are that it works in feature space rather than in data space, incorporates nonlinearities into the loss-function, and integrates information across the entire image. Experiments show better qualitative superpixel segmentation using our methodology than SLIC.

\begin{figure*}[ht!]
\includegraphics[width=10cm]{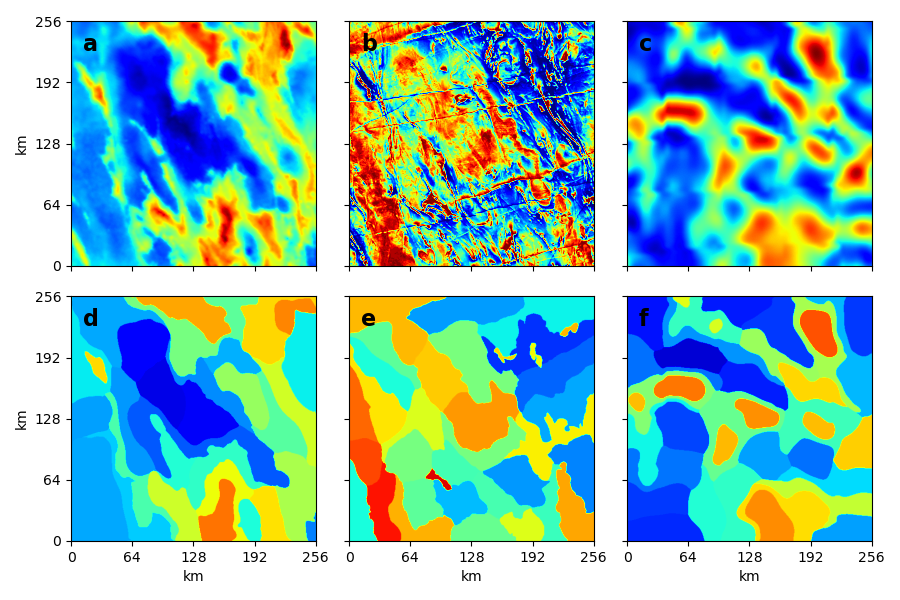} 
\centering
\caption{Superpixel segmentation applied to individual data layers. Columns correspond to gravity, magnetic, and tau data. Target of 35 superpixels per dataset, with realized values of 32, 35, and 39 respectively.  a-c) Data d-f) Corresponding map of super pixel representation ($\mathbf{I}^h$).}
\label{fig:sp_ind}
\end{figure*}

\begin{figure*}[ht!]
\includegraphics[width=\textwidth]{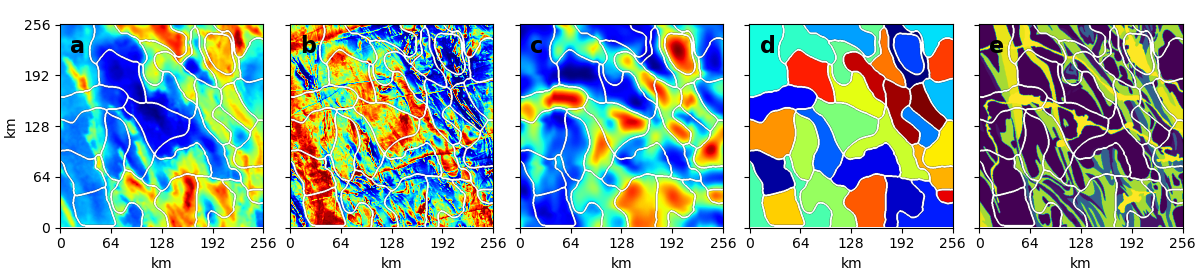}
\centering
\caption{Superpixel segmentation applied to the combination of gravity, magnetic, and tau data layers. (c.f. image with superpixel segmentation applied layer-wise). Targeted and realized number of superpixels is 35 and 33. a-c) Individual gravity, mag, tau layers with combined segmentation overlaid on top. d) Superpixels with random color assignment and outlines in white. e) Superpixels overlaid onto geological map.}
\label{fig:sp_mgd}
\end{figure*}

\begin{figure*}[ht!]
\includegraphics[width=\textwidth]{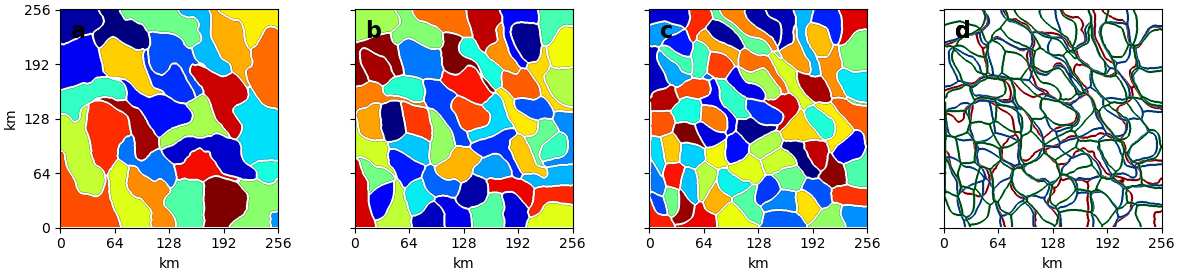}
\centering
\caption{Three superpixel segmentations of the combined magnetic, gravity, and tau datalayers. The target (realized) number of superpixels are a) 30 (30). b) 60 (54). c) 90 (83). Panel d) shows all three segmentations overlayed with the colours red, blue, green corresponding to panels (a), (b), and (c) respectively. }
\label{fig:sp_vparam}
\end{figure*}

Superpixel segmentation (\cref{fig:sp_ind}) is applied independently to gravity, magnetic, and tau datasets in our study domain. Qualitatively we observe good results, with the individual segmentations reflecting the patterns in the data. The best correspondence between the superpixelated image and the original dataset is for the tau image, which has the smoothest features. However, our segmentation performs well for the highly textured magnetic image. Low frequency features are broadly captured (\cref{fig:sp_ind}b), with some high frequency features being retained.

We observe little correlation between the superpixel segmentations, which is unsurprising based on the observation that the images look different and that these datasets provide approximately independent geoscientific information. These provide a baseline for comparison with respect to holistic superpixel generation.

Next we proceed with segmenting the study domain based on all three datasets (\cref{fig:sp_mgd}). We observe that superpixels are generally coherent across all the datasets, although instances where the delineation of a superpixel is arbitrary for a particular layer. Although the aim of superpixels is not to recreate a geological map (nor is it necessarily possible), we compare the superpixel map to a geological map of the region for completeness.

Superpixel segmentation is amenable to a human-in-the-loop workflow. \cref{fig:sp_vparam} shows the results of three different superpixel segmentations applied to the combination of the datalayers with increasing numbers of superpixels. Numerous places exhibit consistent boundaries. This indicates that where the model is well constrained the output is similar, while showing the adaptability to display different solutions where the segmentation may be more complex.

\section{Conclusion} 
\label{sec5}

We present a superpixel approach for fusing geoscientific data into a representative spatial segmentation. This provides an objective, semi-automated method for segmenting a domain into regions with similar physical characteristics. Further, we formulate a new loss function for superpixels, including a novel regularization parameter based on connected-components that allows user-control over the number of superpixels generated.

The method we present is complementary to expert judgement, and in practice can provide a baseline result to be interpreted or modified in the context of aggregate information. Additionally, different solutions can be quantitatively compared using the loss function and parameter choices can be transferred across spatial domains. By handling an arbitrary number of input layers our methodology can be leveraged in cases where qualitative understanding is most difficult.

While we demonstrate our methodology with an application to geophysical data motivated by mineral exploration, it is applicable across a spectrum of geospatial datasets and use-cases. A natural extension beyond the continuous raster data we consider would be to incorporate point data and layers with partial coverage.

{\small
\bibliographystyle{ieee_fullname}
\bibliography{main}
}

\end{document}